# Advantages and a Limitation of Using LEG Nets in a Real Time Problem

Thomas B. Slack
PAR Goverment Systems Corporation
220 Seneca Turnpike
New Hartford, NY 13413

## Abstract

After experimenting with a number of non-probabilistic methods for dealing with uncertainty many researchers reaffirm a preference for probability methods (Lemmer 1977) (Cheeseman 1985), although this remains controversial. The imporatance of being able to form decisions from incomplete data in diagnostic problems has highlighted probabilistic methods (Lemmer 1983) which compute posterior probabilities from prior distributions in a way similar to Bayes Rule, and thus are called Bayesian methods. This paper documents the use of a Bayesian method in a real time problem which is similar to medical diagnosis in that there is a need to form decisions and take some action without complete knowledge of conditions in the problem domain. This particular method has a limitation which is discussed.

## Background and Scope

It has been shown (Lemmer and Barth 1982) (Lemmer 1983) that it is not necessary to form a prior distribution over the joint events of all variables in a given event space to produce a posterior probability for a given event; Component Marginal Distributions (CMDs) can be formed over the joint events of smaller related groups of variables called Local Event Groups (LEGs). Each LEG is related to other LEGs by common variables called intersection sets. The LEGs thus form a network of CMDs, related by their intersection sets, called a LEG Net. By specifying known relationships from statistical measurements and estimating higher order effects from expert knowledge (Lemmer and Norton 1986), prior probabilities can be calculated for the CMDs in a LEG Net (subject to some limitations on the structure of the Net) which are consistent with an assumed underlying distribution over all variables in the LEG Net. This estimation technique makes minimum information assumptions when constraints on a distribution are not fully specified. Once consistent distributions are made, updating techniques (Lemmer and Barth 1982) reflecting the introduction of data for known values of variables in a specific case, guarantee continued consistency of the CMDs posterior to the introduction of the data. The new distributions represent a minimum information update of the prior distributions (ibid), and the values of the unspecified variables represent posterior probabilities given the introduced data.

Several knowledge engineering tools have been developed at PAR Government Systems Corporation consistent with the LEG Net method of dealing with uncertainty. These tools provide the knowledge engineer with the ability to specify LEG Nets and CMDs in each LEG. They also provide the ability to test scenarios within the LEG Net, so specified, to allow a level of confidence in the resultant system (Barth and Norton 1986). There is even a tool which allows explanation of inferencing done by the system (Norton 1986). These tools use *binary evidence variables* (evidence variables which have values only true or false) for all evidence input to the system as a matter of convenience (Barth and Norton 1986) (Norton 1986). The techniques of updating seem to hold the promise that uncertain evidence, may be represented by evidence variables with a domain of a continuous interval. Consistent distributions of joint relationships between variables can be generated by varying any variable in the system and propagating the result throughout the LEG Net. The use of evidence that is not binary is introduced and a limitation of this technique when using such evidence is discussed below.

## The Problem Specification

The problem is a form of real time pattern classification. The classification space is binary, i. e., we simply wish to decide whether this particular collection of input data, called an object, represents something important enough to note, to be marked Yes (or Y), or whether it does not, to be marked No (or N). Objects may also be classified Unknown (U). Performance measures will only reward algorithms that correctly identify objects Y or N, and penalize algorithms that incorrectly identify objects Y or N.

More than one object may occur within a given interval of time called a slot. Slots are equal intervals of time on the order of tenths of a second in length. In each consecutive slot, new objects may occur that were not in the old one, objects may occur again that correlate with previous ones, and objects may disappear in successive slots. It is assumed that other processing will handle the correlation of objects from slot to slot, and the problem at hand is only to classify the current set of objects for the current slot in real time. Although the history of a given object can and should be taken into consideration when classifying, it should be remembered that the nature of the correlation process itself is subject to error, and therefore credibility can not always be given to the object's history.

Many tests of various quality have been devised to relate the different properties of a given object to a Yes or No classification. These tests return values that may be normalized to a probability that the object is a Yes. Such tests correspond to tests that a physician might use to help in medical diagnosis, the difference being that this system must rediagnose each current object at each slot and that the tests are generally nondestructive. The tests also represent Evidence Variables in our LEG Net that are not Binary. The only costs involved with the tests themselves are the costs in time to execute them. The tests are applicable to the problem only under conditions. Tests will be executed for each object within each slot in which the conditions of applicability are met. The algorithms for combining the probabilities must be codable in FORTRAN[1] and must run in real time on a current micro-computer with hardware floating point support.

## The Implemented Solutions

Two solutions to the combination of probabilities in the real time classification problem from the last section are discussed below: In one, a first order Markov model, is used to form the total probability at each slot from the individual tests; in the other, a LEG Net is used. With both methods of combining data, the integration of probability over time is handled by a Markov model. The difference being that the simple Markov Chain method uses the same calculation across the different tests as it does from one time slot to the next, and the Bayesian method integrates each test over time using the Markov Chain, and then combines the current probabilities of each test at each time slot into one total probability using a LEG Net.

---

1. The final version may be machine language or a combination of other high level languages, such as ADA, and machine code; however, in order to be compatible with system simulation efforts, the algorithms to be tested were coded in FORTRAN. Although the FORTRAN version did not have to operate in real time, it needed to be quick enough to allow reasonable turnaround in performance testing.

The output of each test is two probabilities for each object.
1. P(Y'|N) The probability from the current data of the object that it is now a Yes given that it really is a No (Probability of noise).
2. P(N'|Y) The probability from the current data of this object that it is now a No given that really is a Yes (Probability of miss altogether).

From these the P(Y'|Y) = 1 - P(N'|Y) and P(N'|N) = 1-P(Y'|N) are calculated, and the new values of P(N') and P(Y') are obtained using the iterative equation (Equation 1) and the old values of P(N) and P(Y).

$$\begin{bmatrix} P(N') \\ P(Y') \end{bmatrix} = \begin{bmatrix} P(N'|N) & P(N'|Y) \\ P(Y'|N) & P(Y'|Y) \end{bmatrix} * \begin{bmatrix} P(N) \\ P(Y) \end{bmatrix} \quad (1)$$

Each test in turn is allowed to modify the total probability using Equation 1 as if it were the only test being used. After the sequence of tests, the values of P(N) and P(Y) then contain their new values for this slot. This method of updating is subject to error as it makes the assumption that a given test knows that it is following another one, so that the total so far has been thus conditioned by it. The assumption is that it has adjusted its probabilities to account for the fact that the previous tests just occured. In actuality, no such information is used to form the tests. This source of error is aggravated by the conditional inclusion or exclusion of tests. In order to reduce this error, extensive scenario simulation is done and the order of tests is changed to have the best discriminator appear last. The relative strengths of probabilities are adjusted incrementally so that a level performance is obtained. This is called tuning the classifier.

The LEG Net solution integrates the tests over time for each test separately in the same manner as the Markov Chain, then it combines the resultant P(Y)s from each test at each time into one P(Y) for each object. In order to simplify the FORTRAN solution, alll LEGs were assummed to contain three variables, two input variables and one output variable. The CMDs are each $2^3$ or eight numbers corresponding to the possible joint events for the variables in each LEG, as shown in Figure 1.

The constraint of one output variable for each LEG limits the intersection sets of LEGs to one variable each. The updating algorithm is then simplified to multiplying each joint probability that contains a modified event probability by the ratio of its new to old value, and multiplying the others by the ration of (1-new) to (1-old) values. For example, if I1 changes the distribution changes according to Equations 2 and 3.

The specification of the prior values in each LEG was done in LISP using estimation techniques (Lemmer and Norton 1986). The resulting distributions were then coded into the FORTRAN arrays for simulation.

```
O  I2 I1
0  0  0   P0
0  0  1   P1
0  1  0   P2
0  1  1   P3
1  0  0   P4
1  0  1   P5
1  1  0   P6
1  1  1   P7
```

$P(I1) = \sum Pn; n \in \{1\ 3\ 5\ 7\}$

$P(I2) = \sum Pn; n \in \{2\ 3\ 6\ 7\}$

$P(O) = \sum Pn; n \in \{4\ 5\ 6\ 7\}$

**Figure 1**

For $n \in \{1\ 3\ 5\ 7\}$:

$$P'n = Pn * \frac{P'(I1)}{P(I1)} \tag{2}$$

For $n \in \{0\ 2\ 4\ 6\}$:

$$P'n = Pn * \frac{[1-P'(I1)]}{[1-P(I1)]} \tag{3}$$

## Advantages

The LEG Net solution has advantages over the Markov Chain method of combining probabilities. The Leg Net may encode knowledge from an understanding of the conditions under which a test gives good results into the structure of the net. The Markov Chain method may only encode such knowledge as a "go" or "no go" applicability of such a test. Another advantage is that knowledge gained by experience of using the tests can be used to change directly the relative strengths of each test, and its effect on the goal variable. The effect of a variable on the whole LEG Net is shown by the change of the distribution. The Markov Chain method only allows strengthening or weakening of the tests themselves, or a changing of the order of the tests. The effect of changing the order in the Markov Chain is dependent on the relative strengths of the tests themselves and it is therefore difficult to show whether a given test actually contributes to the final probability of a given object within a slot, or whether its effect is being masked by a test later in the chain. These advantages can significantly reduce the time needed to tune the performance of the classifier during system testing. It also allows easier changes with the introduction of new test data.

The LEG Net, of course represents advantages in size and speed over using Bayes Rule itself. If we had enough room to specify the entire distribution for our problem (11 tests + 1 goal, 2^12 or 4K position array of point numbers), and could gather enough statistical information to populate the distribution, and had the computational power to compute Bayes Rule for that distribution, then we could use it and compute the actual distribution for a given set of values of EVs. The LEG Net gets a consistent distribution (although not quite the same one) without that size and speed disadvantage.

## Limitation

When this project was started, it was hoped that an advantage of using a LEG Net would be that the order of introducing data would not have an effect on the result.

# ADVANTAGES AND LIMITATIONS TO USING A LEG NET FOR A REAL TIME PROBLEM

A Markov Chain certainly is order-dependent, and it would be useful if we could (as when using a strict Bayes Rule approach) apply the changes in the variables in any order. Indeed this is the case when binary evidence variables are used to update a LEG Net. Such evidence conditions the distribution as much as is possible when it is introduced, and can therefore no longer affect or be affected by it; binary evidence always introduces zeros into the joint distribution, and such zeros are unaffected by further updating by multiplying ratios.

It was discovered, however, that the LEG Net also contains order dependency when non-binary evidence is introduced. As each piece of evidence is incorporated into the network, the distribution changes. Each new posterior distribution becomes the prior distribution for the next piece of evidence. Unless two variables are independent (as determined by the prior distribution) changing one will change the other. The earlier variables values will be changed by the later ones.

This observation should not really have been a surprising one; it is mentioned in the 1985 uncertainty workshop (Vaugham et al 1985). In fact, it is recognized in the previous papers on Generalized Bayesian updating using LEG Nets (Lemmer and Barth 1982) (Barth and Norton 1986) by their examples of LEG Nets. If two LEGs in a LEG Net have two variables in their intersection set, then when one LEG changes, the updating algorithm is much more complicated than the one given above. Essentially, a distribution is formed from the changed leg over the joint probabilities of events in the intersection set. This distribution is then used to update the other LEG instead of updating each variable separately. This is necessary so that the updated values for both posterior probabilities be considered together when the LEG CMD is updated. What is not clear is that the same kind of calculation is needed whenever two non binary evidence variables which are not in the same LEG are updated. If the change in two variables from different LEGs in a LEG Net represents a coincident change in the underlying distribution, then they must be considered at the same time when updating the net. As in a Markov Chain, the order dependency of LEG Nets is that the last is the most important. As each test result is added to the net, the last one updated will have the actual value given while the first one is the most suseptable to being modified.

The importance of the order dependency of variables in LEG Nets in general can be understood by answering three questions.

1. How much does the order dependency affect results?
2. How can the limitation be avoided?
3. How can the limitation be "solved" and the correct CMDs be calculated?

Suppose we wish to find a distribution consistent with the prior in Figure 2 but with $P(I1) = 0.9$ and $P(I2) = 0.1$. If we choose the order I1 first and then I2, Table 1(a) summarizes the calculations. If we wish to have the same distribution, but choose order I2 first and then I1, then table 1(b) summarizes the calculations. Note the difference in P(O) depending on the order of updating the variables (0.5814776 as opposed to 0.6673082).

In order to avoid the order dependency, it should be remembered that: First, the order dependancy does not occur unless non binary evidence variables are used; if it is possible to only use BEVs then this should be done. However, in the case of tests to be performed which have qualitative values statistically correlated to the goal variable this may not be a viable solution. Second, since the effects of variables on one another are a direct result of dependence between input variables,

425

Prior Distribution (joint probabilities)    Prior event probabilities

| O | I2 | I1 |      |
|---|----|----|------|
| 0 | 0  | 0  | 0.40 |
| 0 | 0  | 1  | 0.05 |
| 0 | 1  | 0  | 0.05 |
| 0 | 1  | 1  | 0.00 |
| 1 | 0  | 0  | 0.00 |
| 1 | 0  | 1  | 0.10 |
| 1 | 1  | 0  | 0.10 |
| 1 | 1  | 1  | 0.30 |

$P(I1) = 0.45$
$P(I2) = 0.45$
$P(O) = 0.50$

**Figure 2**

| Variable | P(I1)     | P(I2)     | P(O)      | Error     |
|----------|-----------|-----------|-----------|-----------|
| I1       | 0.9000000 | 0.6272727 | 0.8181819 | 0.5272727 |
| I2       | 0.8200424 | 0.1000000 | 0.5814776 | 0.0799575 |

(a)

| Variable | P(I1)     | P(I2)     | P(O)      | Error     |
|----------|-----------|-----------|-----------|-----------|
| I2       | 0.3121212 | 0.1000000 | 0.2525252 | 0.5878788 |
| I1       | 0.9000000 | 0.1970188 | 0.6673082 | 0.097078  |

(b)

**Table 1**

if tests are constructed such that their probabilities are independent, there will be no order dependence. This is often not an acceptable solution as one major purpose in using a LEG Net is to avoid making independence assumptions.

We have found that by iterativly operating on a LEG Net, we can continue to update the several variables in turn until the values converge to the desired distribution. Continuing with the calculations from Table 1, Table 2 has two more iterations for each order of evaluating the inputs.

Another effort needs to be made to determine under what conditions such a network will converge to a consistent distribution with the values of each evidence variable correct and when it will not. The fact that the values converge in our case allows us to completely ignore the limitation because the tests are oversampling the input data. This means that the values from the tests vary slowly from slot to slot and therefore if we use the posteriors from one slot as the prior for the next, they will reach approximately the same answers as would be given by Bayes rule delayed by the settling time of the LEG Net. We may reduce our settling times by increasing our sampling rate, subject to the limits of the cost of calculation in the Net. A different restriction arises if we use LEG Nets by allowing the posteriors to become the new priors. We should never allow the values of the evidence to reach either 0 or 1. If they do, they cannot change further. Therefore, using the LEG Net in this fashion the requires the use of evidence variables that are not binary as opposed to being required to use binary ones.

A solution to the problems arising out of the order dependency of variables in LEG Nets, which does not depend on the convergence of the distribution, requires a way of fixing a value in the net that is not a one or zero; a way of declaring that this

| Iteration | Variable | P(I1) | P(I2) | P(O) | Error |
|---|---|---|---|---|---|
| 1 | I1 | 0.9000000 | 0.6272728 | 0.8181819 | 0.5272728 |
|   | I2 | 0.8200424 | 0.1000000 | 0.5814776 | 0.0799575 |
| 2 | I1 | 0.9000000 | 0.1073947 | 0.6366036 | 0.0073947 |
|   | I2 | 0.8993579 | 0.1000000 | 0.6336551 | 0.0006421 |
| 3 | I1 | 0.9000000 | 0.1000554 | 0.6340969 | 0.0000554 |
|   | I2 | 0.8999953 | 0.1000000 | 0.6340749 | 0.0000047 |

(a)

| Iteration | Variable | P(I1) | P(I2) | P(O) | Error |
|---|---|---|---|---|---|
| 1 | I2 | 0.3121212 | 0.1000000 | 0.2525252 | 0.5878788 |
|   | I1 | 0.9000000 | 0.1970188 | 0.6673082 | 0.0970788 |
| 2 | I2 | 0.8908822 | 0.1000000 | 0.6280743 | 0.0091178 |
|   | I1 | 0.9000000 | 0.1007928 | 0.6343487 | 0.0007928 |
| 3 | I2 | 0.8999316 | 0.1000000 | 0.6340329 | 0.0000684 |
|   | I1 | 0.9000000 | 0.1000059 | 0.6340800 | 0.0000059 |

(b)

**Table 2**

value is posterior and should not change. One can envision this process to be similar to the process of the estimation of priors, and would entail the propagation of intervals of values that constrain variables throughout the net. Further research is needed to expose such methods.

## Performance

The Actual LEG Net used and its performance can not be presented here. Performance was measured by a Monte Carlo simulation of input data. The measured parameters of each object were varied assuming a gaussian distribution. A total of 500 data points was generated for each of several scenarios. The points were then passed into each of the tests mentioned above in the statement of the problem. The scores resulting from the tests were then combined both by the Markov model technique and the LEG Net technique. A performance improvement was experienced overall by using the LEG Net technique when the same tests were used for each scenario. When the tests were tuned to match the individual scenarios, it was discovered that for each scenario, the Markov Chain technique could be tuned to match the performance of the LEG Net.

## Summary

It was discovered that there are many advantages to using a LEG Net for the described real time classification problem. In our LEG Net solution, the order dependency was more controllable than that seen in the Markov Chain because of the ability to see the effect of order dependency. The flexibility of changing tests and controlling the effect on the resultant probability demonstrate the effectiveness of LEG nets over the Markov technique. It has been demonstrated that the LEG Net solution can improve performance over the Markov method given that the variability in the scenarios is sufficient not to allow the tuning of the Markov Chain to the data. Costs in execution time are higher than the Markov method but significantly better than the strict use of Bayes rule with comparable results. When using a LEG Net for real time purposes, it becomes necessary to oversample the input data so that it will be quasi static; this allows time for the settling of the order dependency out of the LEG Net. Future work is needed to uncover methods of determining the stability and conditions of convergence of LEG Net distributions, and to discover ways of propagating probabilities through the network that hold designated inputs constant (coincident posterior values).